\documentclass[runningheads]{llncs}
\pdfoutput=1 
\usepackage{graphicx}
\usepackage{subfigure}

\usepackage{tikz}
\usepackage{comment}
\usepackage{amsmath,amssymb} % define this before the line numbering.
\usepackage{color}
\usepackage{multirow}
\usepackage{makecell}
\usepackage[width=122mm,left=12mm,paperwidth=146mm,height=193mm,top=12mm,paperheight=217mm]{geometry}

\begin{document}
	% \linenumbers
	\pagestyle{headings}
	\mainmatter
	\def\ECCVSubNumber{100}  % Insert your submission number here
	
	\title{IKEA Object State Dataset: A 6DoF object pose estimation dataset and benchmark for multi-state assembly objects}

	\author{Yongzhi Su\inst{1,2} \and
		Mingxin Liu\inst{1} \and
		Jason Rambach\inst{2} \and Antonia Pehrson\inst{3} \and Anton Berg\inst{3} \and
		Didier Stricker \inst{1,2}}
	\institute{TU Kaiserslautern, Kaiserslautern, Germany \and
		German Research Center for Artificial Intelligence (DFKI), Kaiserslautern, Germany \and IKEA Marketing \& Communication AB\\
		\email{\{Yongzhi.Su, Mingxin.Liu, Jason.Rambach, Didier.Stricker\}@dfki.de} \\
		\email{\{antonia.pehrson2, anton.berg\}@inter.ikea.com}}
	%\institute{Paper ID \ECCVSubNumber}
	%\end{comment}
	
	\maketitle
	
	\begin{abstract}
		Utilizing 6DoF(Degrees of Freedom) pose information of an object and its components is critical for object state detection tasks. We present IKEA Object State Dataset, a new dataset that contains IKEA furniture 3D models, RGBD video of the assembly process, the 6DoF pose of furniture parts and their bounding box. The proposed dataset will be available at \textit{\url{https://github.com/mxllmx/IKEAObjectStateDataset}}.
	\end{abstract}

	\section{Introduction}
	In recent years, AR (Augmented Reality) has been one of the most popular technologies in Industry 4.0 and intelligent manufacturing. The use of AR is beneficial for improving and accelerating product and process development. One application of AR is maintenance-assembly-repair. AR assisted assembly is proved significantly faster than manual assembly under paper-based or digital tablet instructions~\cite{funk2016interactive}. It can help operators to understand the presented information, to shorten the production time and to avoid human errors. Such assembly tasks need to deal with objects that consist of several removable and adjustable components, so object state detection, i.e. detecting the current assembly step or state of objects, is very important for related application development.
	
	Existing object state detectors~\cite{su2019deep,liu2020tga} estimate the 6DoF pose of each object's component, then determine the current assembly step. They rely on the deep learning method thus need large datasets with 6DoF pose information for training. Considering realistic conditions such as time and hardware, using large-scale synthetic data for training and  small portion of real data for testing is a compromised method. However, there is still a reality gap between synthetic data and real-world data. Even though the rendering engine can generate highly photorealistic or generalized synthetic objects datasets, synthesizing human motions during assembly is complex and challenging. In a real assembly scene, the operator may interact with the objects in varied and unexpected ways to achieve the goal. The generated occlusion and possible interactions are difficult to be reflected in a synthetic dataset.
	
	There are assembly datasets~\cite{toyer2017human,ben2021ikea} that combine human action recognition and object information together. Nevertheless, their works focus more on human poses, which is helpful for better understanding and detecting patterns in task-oriented human activities but has limitations on the collection of object locations and orientations. Because only object bounding boxes~\cite{toyer2017human} or object segmentation annotations~\cite{ben2021ikea} are included. Other famous objects 6DoF datasets~\cite{brachmann2014learning,hodan2017t,kaskman2019homebreweddb,xiang2017posecnn} provide  non-deformable rigid objects' position and orientation, which are integral and motionless, and not suitable for an assembly task.
	
	Distinct from previous studies, our IKEA Object State Dataset focuses on the 6DoF pose of each object's component in every state. We use IKEA furniture 3D models published by ~\cite{3dMesh}. For every kind of IKEA furniture presented in our dataset, its parts are almost textureless and of the same colour and shape. Therefore, the appearance information in frames could be misleading. Additionally, during the assembly process, the movement of an object part is sometimes slight. It could be just a little clockwise spin and hard to distinguish. The ambiguity is also caused by inter-parts occlusion, human motion occlusion, change of angle, etc. We give our best effort to shoot the video sequence without blind spots, so we set up 4 RGBD cameras with different viewpoints and records the process of IKEA furniture assembly synchronously. Our dataset contains 3D mesh models of the IKEA furniture objects, depth, RGB images and the 6DoF object poses for each frame.
	% complete with annotated 6DOF poses for(for details, see Section 3). Also provided are 3D mesh models of the objects, which may be used for training of recognition algorithms. 
	%The code for utilizing and integrating the dataset with different algorithms is also publicly available.
	%Our contributions are twofold:
	
	\section{Related Work}
	\subsection{6D Object Pose Datasests}
	Many researchers created 6DoF object pose datasets as standard benchmarks for various estimation tasks. They continue to provide innovative thinking and methods for determining the 3D models' position and rotation information in the images. Below are the details of several well-known datasets:
	
	\noindent{\bf LM-O (Linemod-Occluded)~\cite{brachmann2014learning}}: the dataset is improved based on the Linemod dataset~\cite{hinterstoisser2012model}. Instead of using a template-based technique, it uses a decision forest that jointly predicts both 3D object coordinates and objects instance probabilities. This dataset contains 10k images of 20 textured and textureless objects, captured under three different lighting conditions.
	
	\noindent{\bf T-LESS~\cite{hodan2017t}}: this work places emphasis on the textureless rigid objects. The dataset contains 30 industry-relevant objects, which have no significant texture, discriminative colour or distinctive reflectance properties. Their shapes and/or sizes often bear similarities, and a few objects are served as compositions of other objects. Then the result images are divided into $\sim$39K training with a black background and $\sim$10K test images in 20 different scenes.
	
	\noindent{\bf YCB-Video~\cite{xiang2017posecnn}}: it contains 6D poses of 21 YCB objects in 92 videos with a total of 133,827 frames. The household objects own different symmetries. They are arranged in various poses and spatial configurations, and some of the frames also contain severe occlusion between objects.
	
	\noindent{\bf HomebrewedDB~\cite{kaskman2019homebreweddb}}: the dataset is consist of 34,830 images, 33 objects (3 types) over 13 scenes of various difficulty. The scenes' complexity varies regarding the number and size of objects, occlusion and clutter levels. By having a camera pose in each image estimated from the markerboard and having object poses in the markerboard coordinate system, 6D object poses for each of the frames can easily be computed. 
	
	\noindent{\bf Fraunhofer IPA Bin-Picking dataset~\cite{kleeberger2019large}}: both synthetic and real-world scenes are contained in this dataset, which is designed for bin-picking scenarios with robotic arm. The ten objects have both industrial elements like a gear shaft and household objects like a candlestick. For the real data part, the dataset has 520 fully annotated point clouds and corresponding depth images, via fitting point cloud representation and object's CAD model with the Iterative Closest Point (ICP) algorithm. As well as about 206,000 synthetic scenes, the dataset is one of the enormous public datasets for object pose estimation in general.
	
	The overview of these datasets is given in Table~\ref{table:headings}. These works inspired the way of generating a 6DoF object pose dataset. Most of them fix the objects' poses and track objects in the trajectory of cameras. In contrast,  our dataset setup is fixing the cameras, and the objects are movable during the process.
	\begin{table}
		\fontsize{2pt}{3pt}
		\begin{center}
			\caption{Overview of existing 6DoF object pose estimation datasets.}
			\label{table:headings}
			\begin{tabular}{c|ccccc}
				\hline
				\multirow{2}{*}{Dataset}& \multirowcell{2}{Visual \\ Modality} & \multirowcell{2}{Real/Synthetic\\images} & \multirow{2}{*}{\#objects}  & \multirow{2}{*}{resolution} & \multirow{2}{*}{marker}\\ 
				&  &  &  &  &\\ \hline
				Linemod-Occluded~\cite{brachmann2014learning} & RGBD & 10k/- &20&640 x 480& yes\\ \hline
				T-LESS~\cite{hodan2017t}&RGBD& 48k/- & 30 & 1280 x 1024 &yes\\ \hline
				YCB-Video~\cite{xiang2017posecnn}&RGBD& 133827/- & 21 &640 x 480 & no\\ \hline
				\multirow{2}{*}{HomebrewedDB~\cite{kaskman2019homebreweddb}} &\multirow{2}{*}{RGBD}& \multirow{2}{*}{34830/-} & \multirow{2}{*}{30} & \multirowcell{2}{640 x 480 \&\\ 1920 x 1080} &\multirow{2}{*}{yes}\\
				&  &  &  &  &\\  \hline
				Fraunhofer IPA~\cite{kleeberger2019large}& Stereo & 520/206,000 & 10 & 1280×1024 &no\\
				
				\hline
			\end{tabular}
		\end{center}
	\end{table}
	
	\subsection{Assembly Datasets}
	There was IKEA furniture dataset~\cite{lim2013parsing} comprising 800 images and 225 3D models with furniture pose. Another research direction about furniture datasets is furniture parsing. The approach ~\cite{badami20173d} try to find the interaction element (IE) of the furniture item and compute the semantic segmentation by performing subset selection.
	
\begin{figure}
		\centering
		\begin{minipage}[b]{0.16\linewidth}
			\centering
			\includegraphics[width=0.9in]{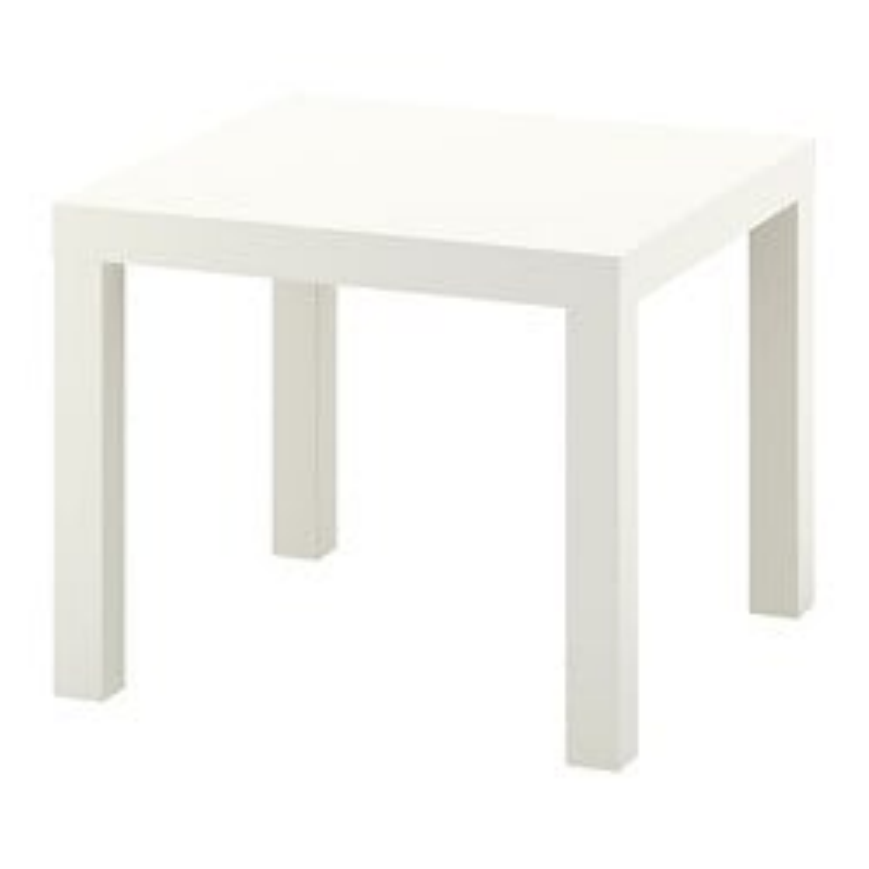}
			\label{fig:minipage1}
		\end{minipage}
		\quad
		\begin{minipage}[b]{0.16\linewidth}
			\centering
			\includegraphics[width=0.9in]{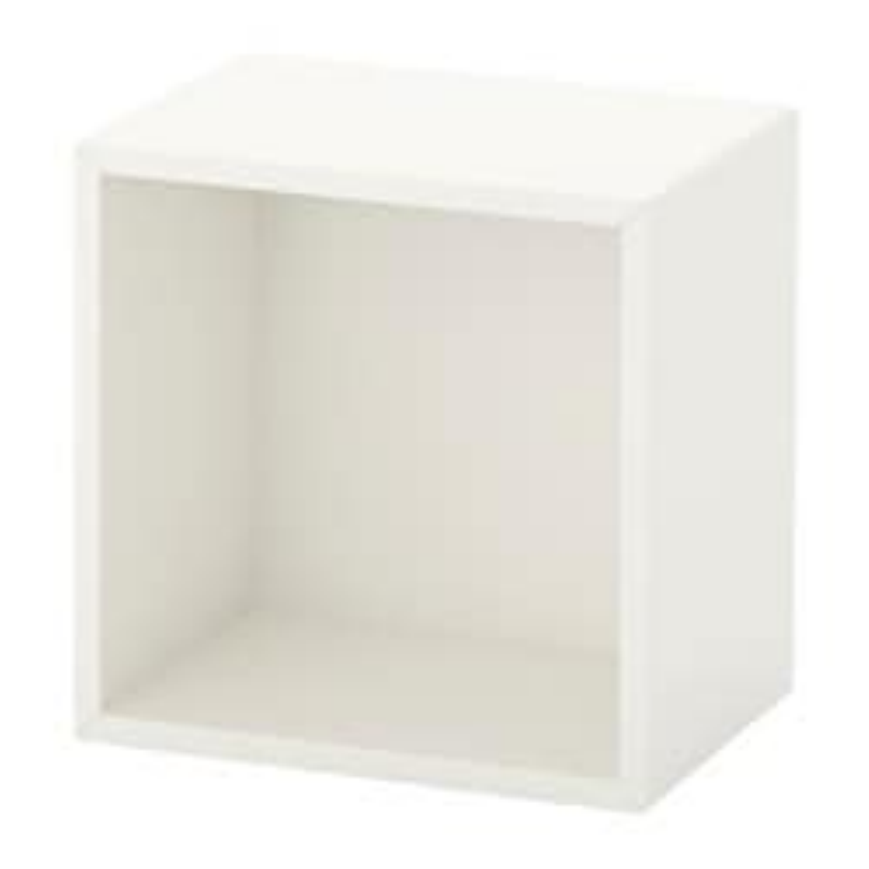}
			\label{fig:minipage2}
		\end{minipage}
		\quad
		\begin{minipage}[b]{0.16\linewidth}
			\centering
			\includegraphics[width=0.9in]{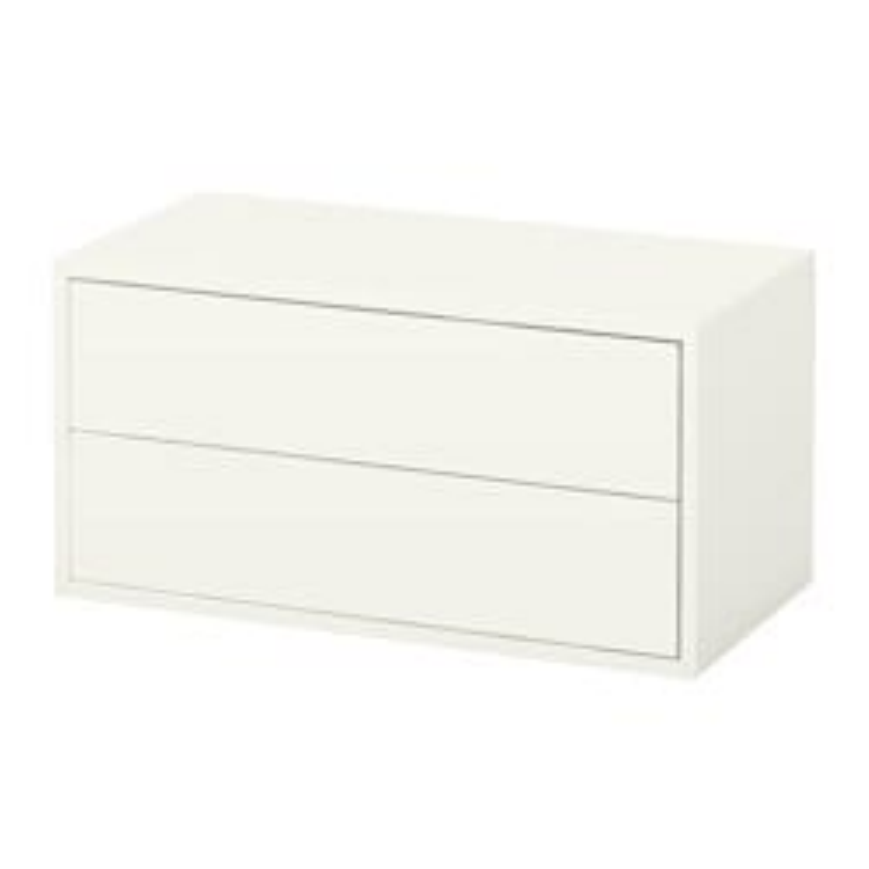}
			\label{fig:minipage3}
		\end{minipage}
		\quad
		\begin{minipage}[b]{0.16\linewidth}
			\centering
			\includegraphics[width=0.9in]{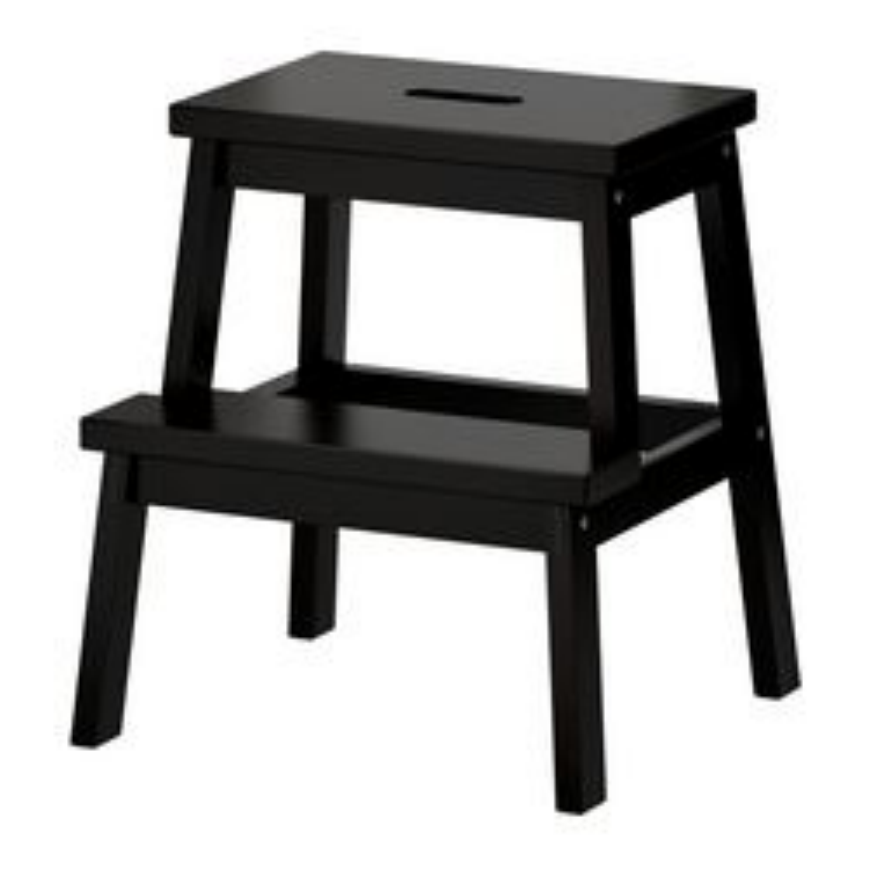}
			\label{fig:minipage4}
		\end{minipage}
		\quad
		\begin{minipage}[b]{0.16\linewidth}
			\centering
			\includegraphics[width=0.9in]{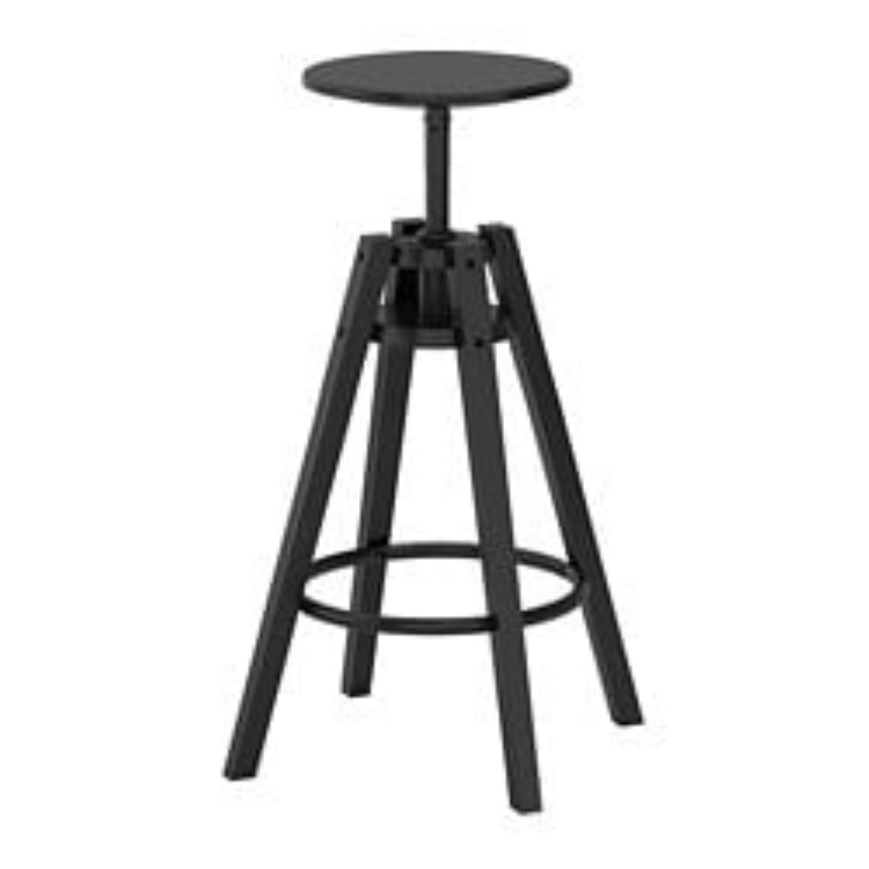}
			\label{fig:minipage5}
		\end{minipage}
		\caption{3D models of IKEA Object State Dataset. Objects from left to right are: a white side table, a white small cabinet, a white cabinet with 2 drawers, a black step stool and a black bar stool.}
		\label{fig:2}
	\end{figure}

	In terms of furniture assembly datasets, Youngkyoon Jang et al.~\cite{jang2019epic} use head-mounted cameras to shoot the process of assembling a camping tent outdoors. However, all 1,171,897 frames are annotated with subtask label, uncertainty label, error label and eye-tracking information, which targets the egocentric task. Chakraborty and Hebert~\cite{chakraborty2021learning} focus on occlusion encountered by object trackers. They try to detect small tools like screws, which are easily occluded, and their dataset is collected from the internet. Similar to our work, Yizhak Ben-Shabat et al.~\cite{ben2021ikea} provide a 3,046,977 frames dataset also with IKEA furniture items. It consists of four different furniture types (side table, coffee table, TV bench, and drawer), 371 assemblies in three views. Each procedure frame is annotated with the human skeleton and the object segmentation that is preferable for human-object interactions research. Our dataset uses similar object models, as depicted in Fig.~\ref{fig:2}, but put effort into object poses.
	
	\section{IKEA Object State Dataset}
	\subsection{Hardware Setup}
	Four Microsoft Kinect Azure cameras are applied in the hardware setup. It is a cutting-edge time-of-flight RGBD camera with sufficient developer tools. We arrange the four cameras in a circle. They are placed in a position higher than the objects, with the lens facing down and aiming at the operating space. The operating space is a classic indoor scene where the operator will assemble the IKEA furniture. Three calibration boards are fixed on the floor to determine the relative pose of the cameras. The computer that we use to run the acquisition program has the following parameters: Intel Xeon(R) CPU E3-1245 v5@3.50GHz x 8, GeForce GTX 1050 Ti/PCle/SSE2 and 64-bit Ubuntu 18.04.6 LTS.
	\subsection{Data acquisition}
\begin{figure}
		\centering
		\begin{minipage}{0.16\textwidth}
			\centering
			\includegraphics[width=0.9in ]{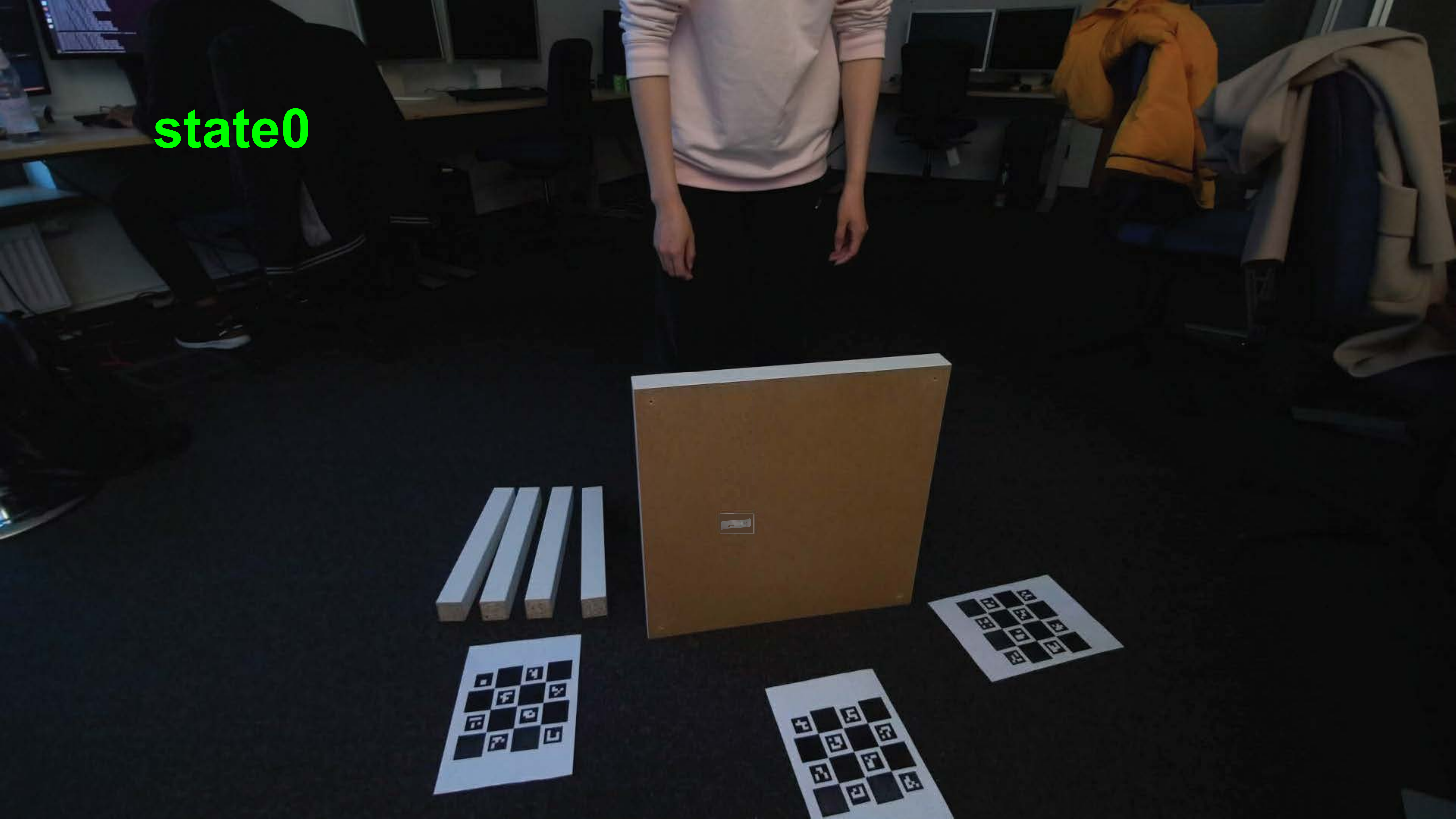}\\
			\includegraphics[width=0.9in ]{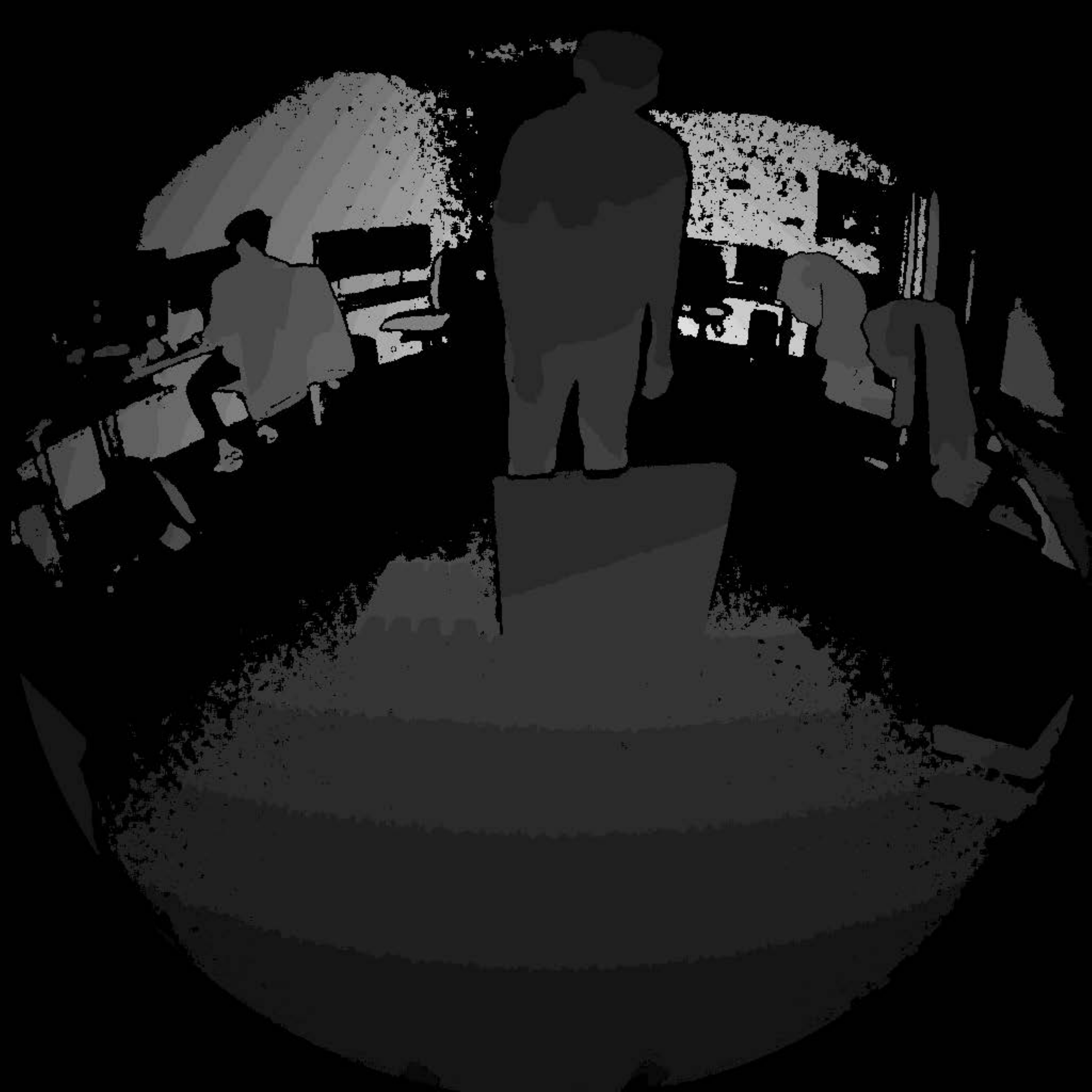}
		\end{minipage}
		\quad
		\begin{minipage}{0.16\linewidth}
			\centering
			\includegraphics[angle=90, width=0.9in ]{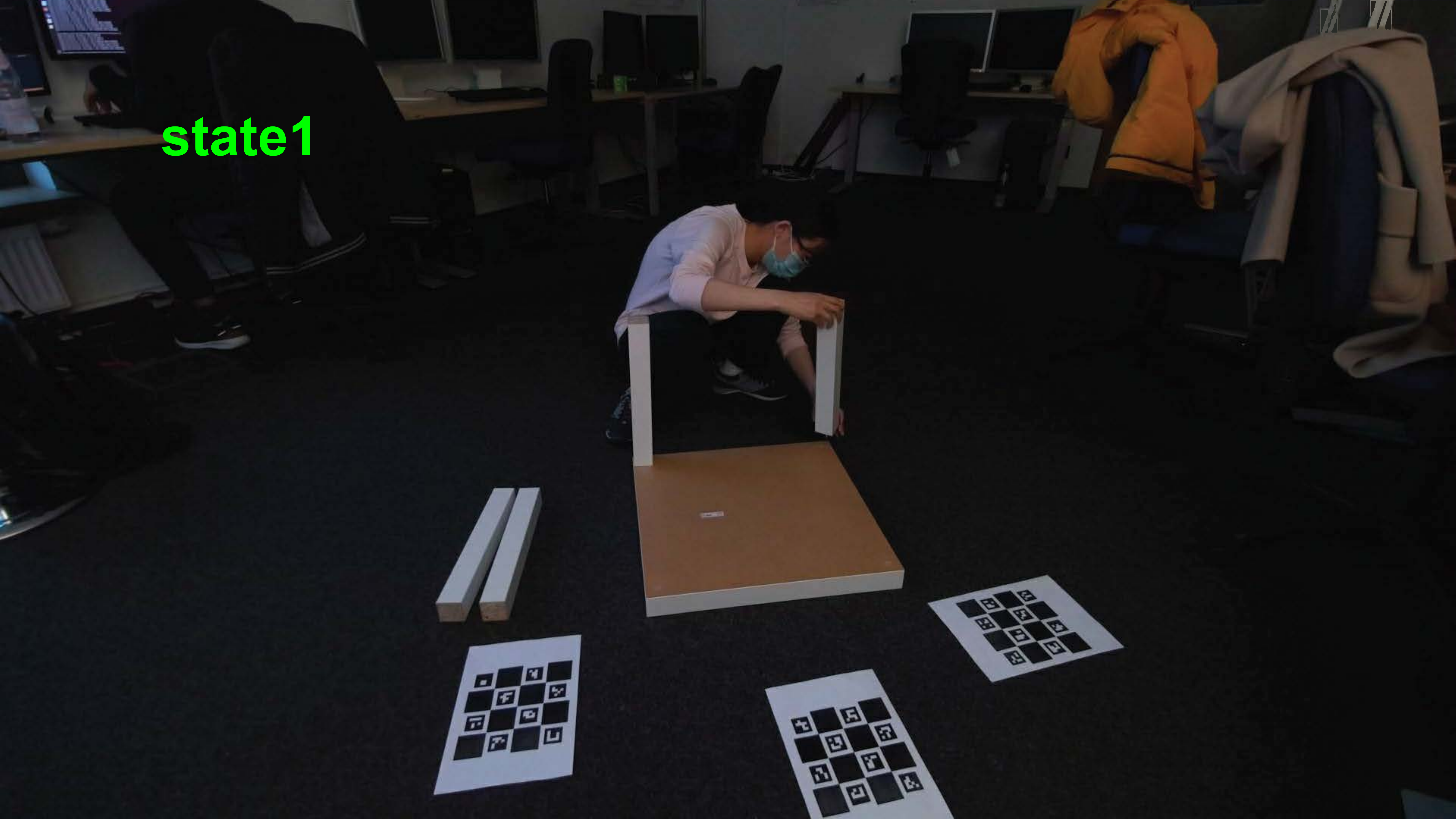}\\
			\includegraphics[width=0.9in ]{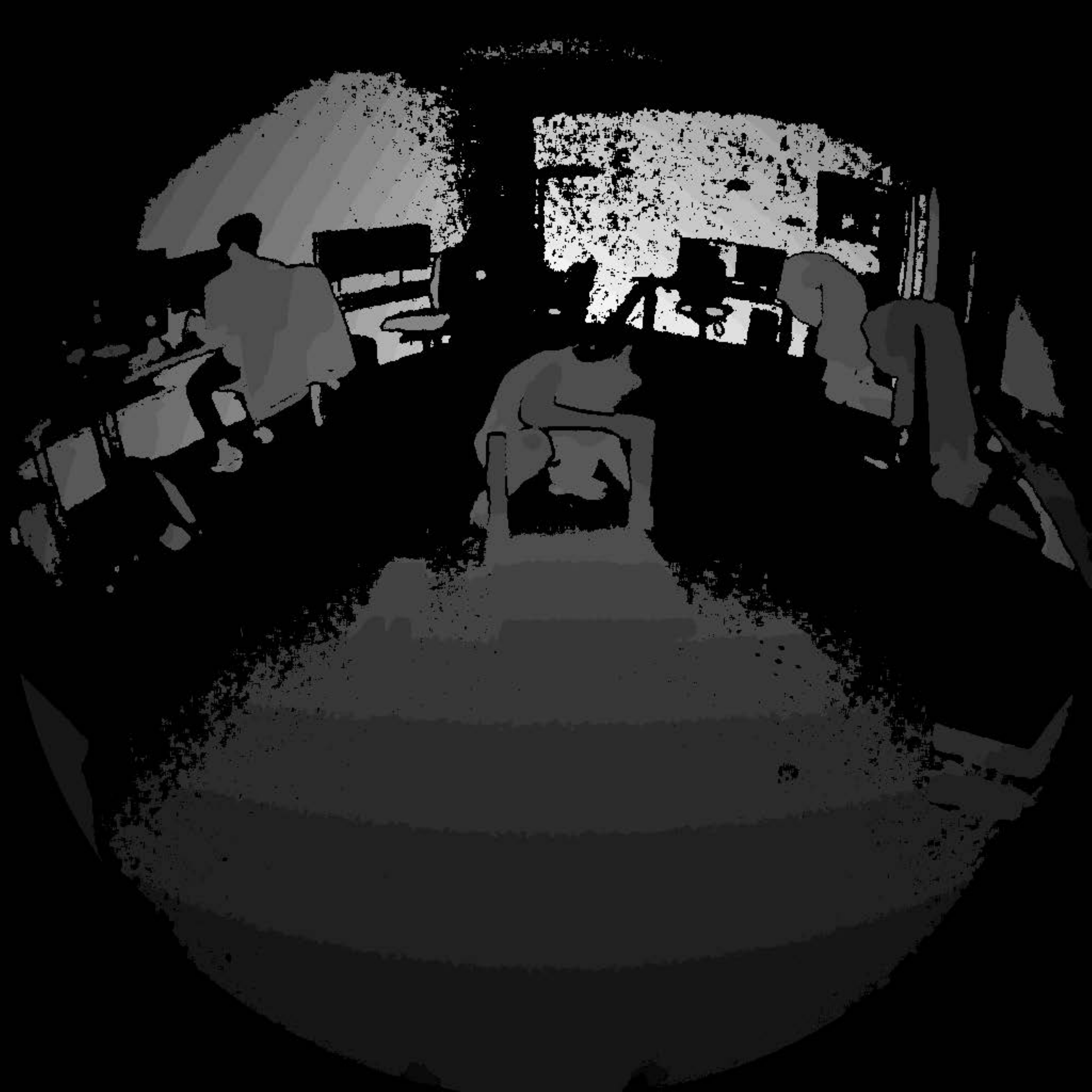}
		\end{minipage}
		\quad
		\begin{minipage}{0.16\linewidth}
			\centering
			\includegraphics[width=0.9in ]{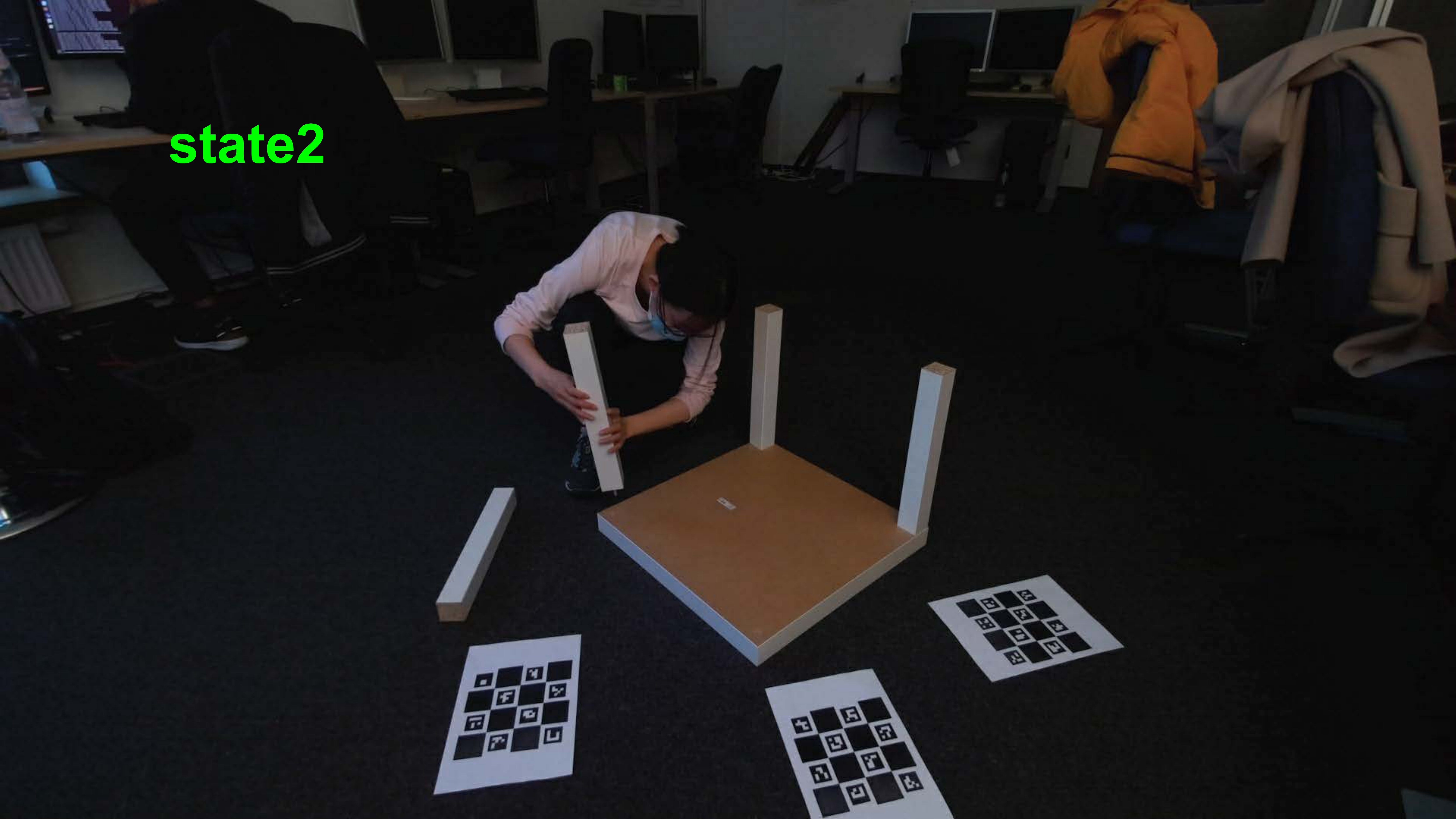}\\
			\includegraphics[width=0.9in ]{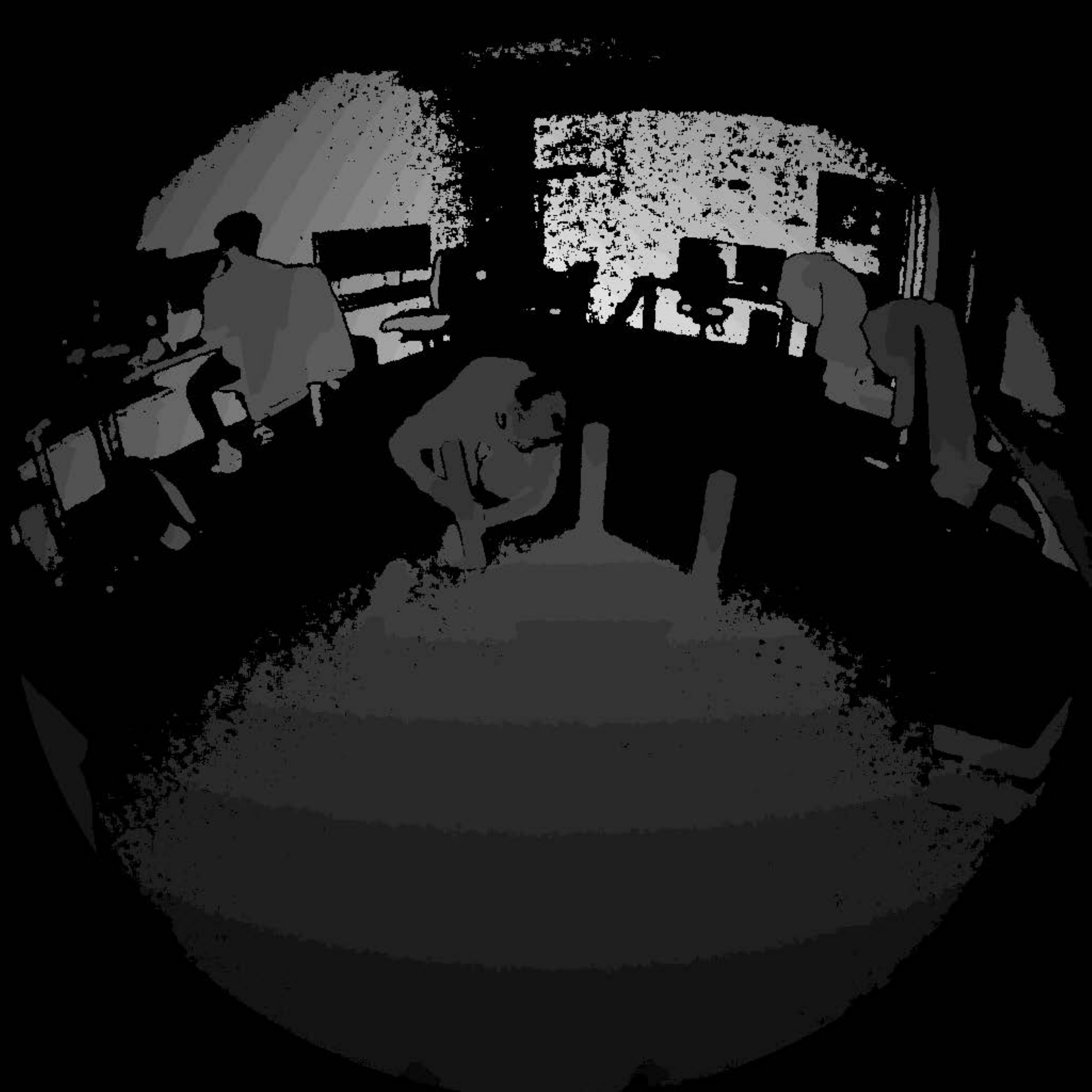}
		\end{minipage}
		\quad
		\begin{minipage}{0.16\linewidth}
			\centering
			\includegraphics[width=0.9in ]{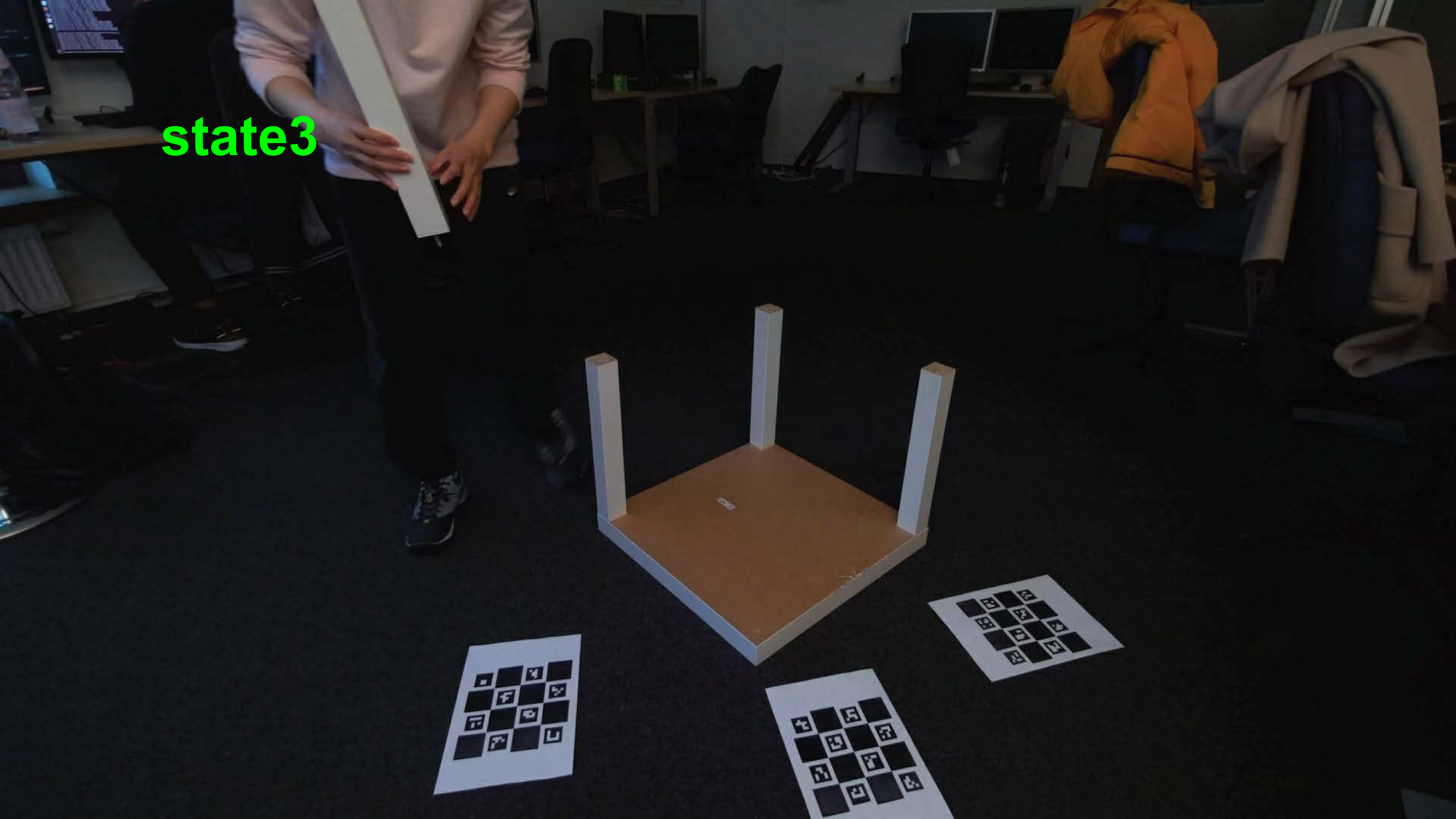}\\
			\includegraphics[width=0.9in ]{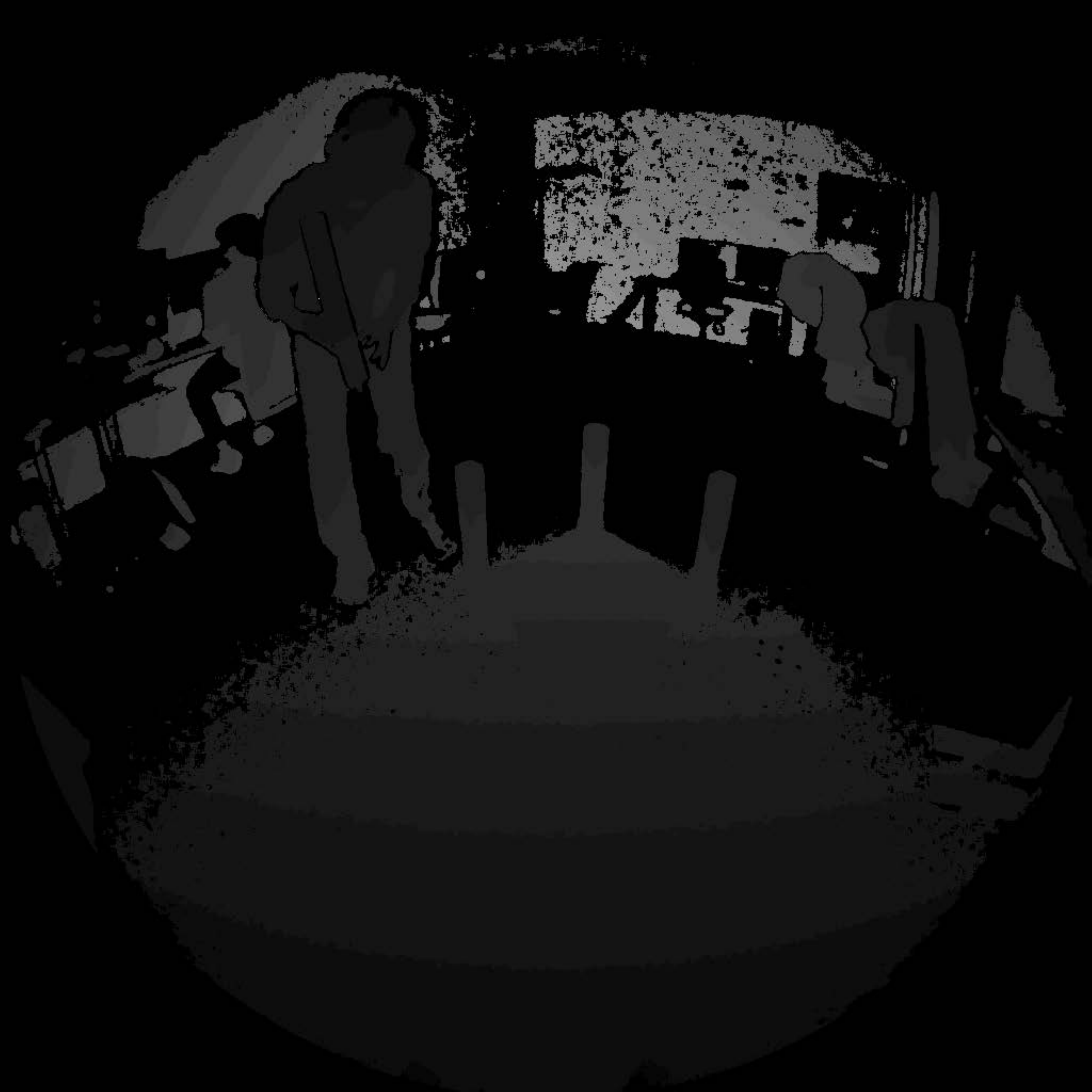}
		\end{minipage}
		\quad
		\begin{minipage}{0.16\linewidth}
			\centering
			\includegraphics[width=0.9in ]{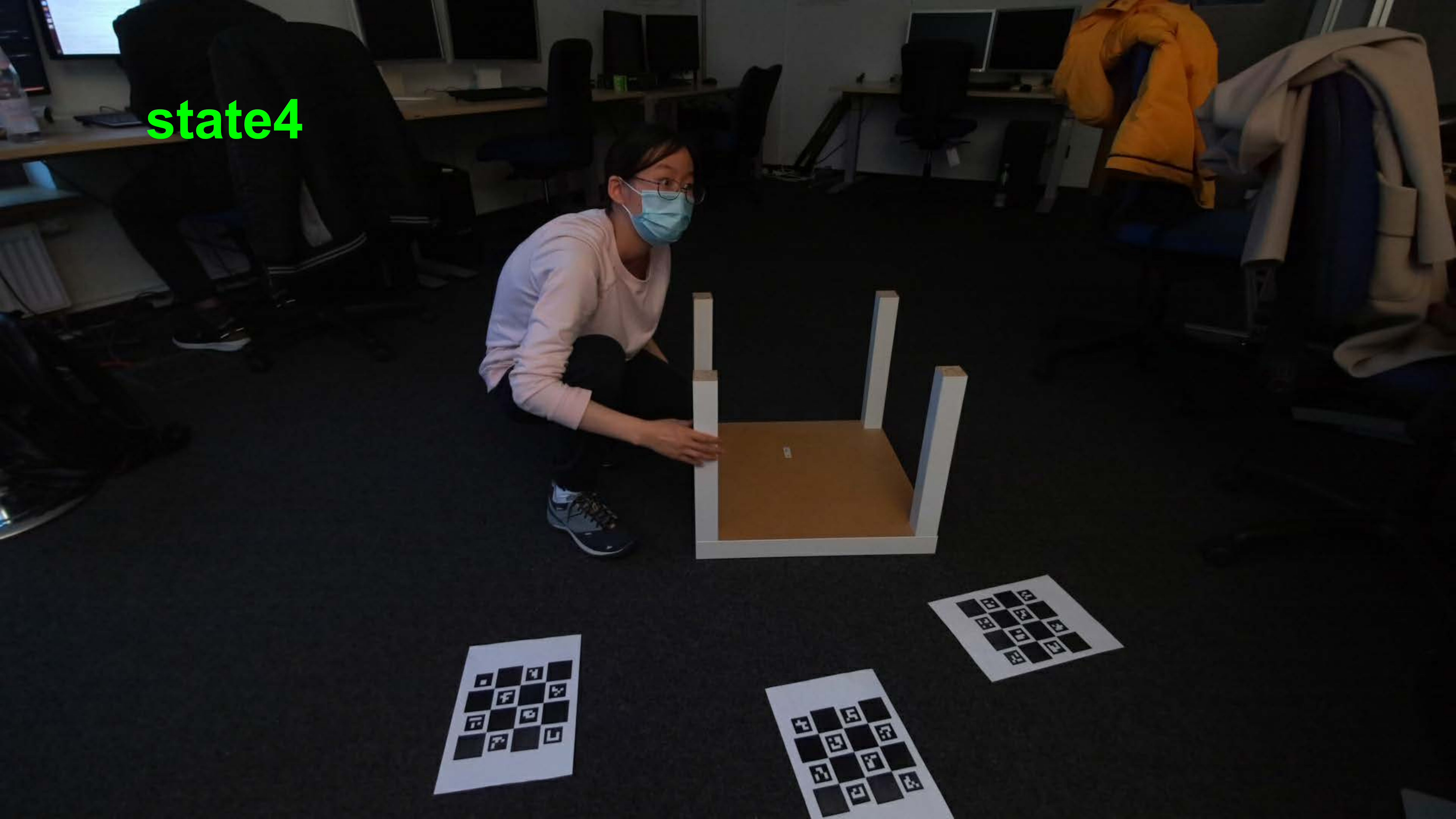}\\
			\includegraphics[width=0.9in ]{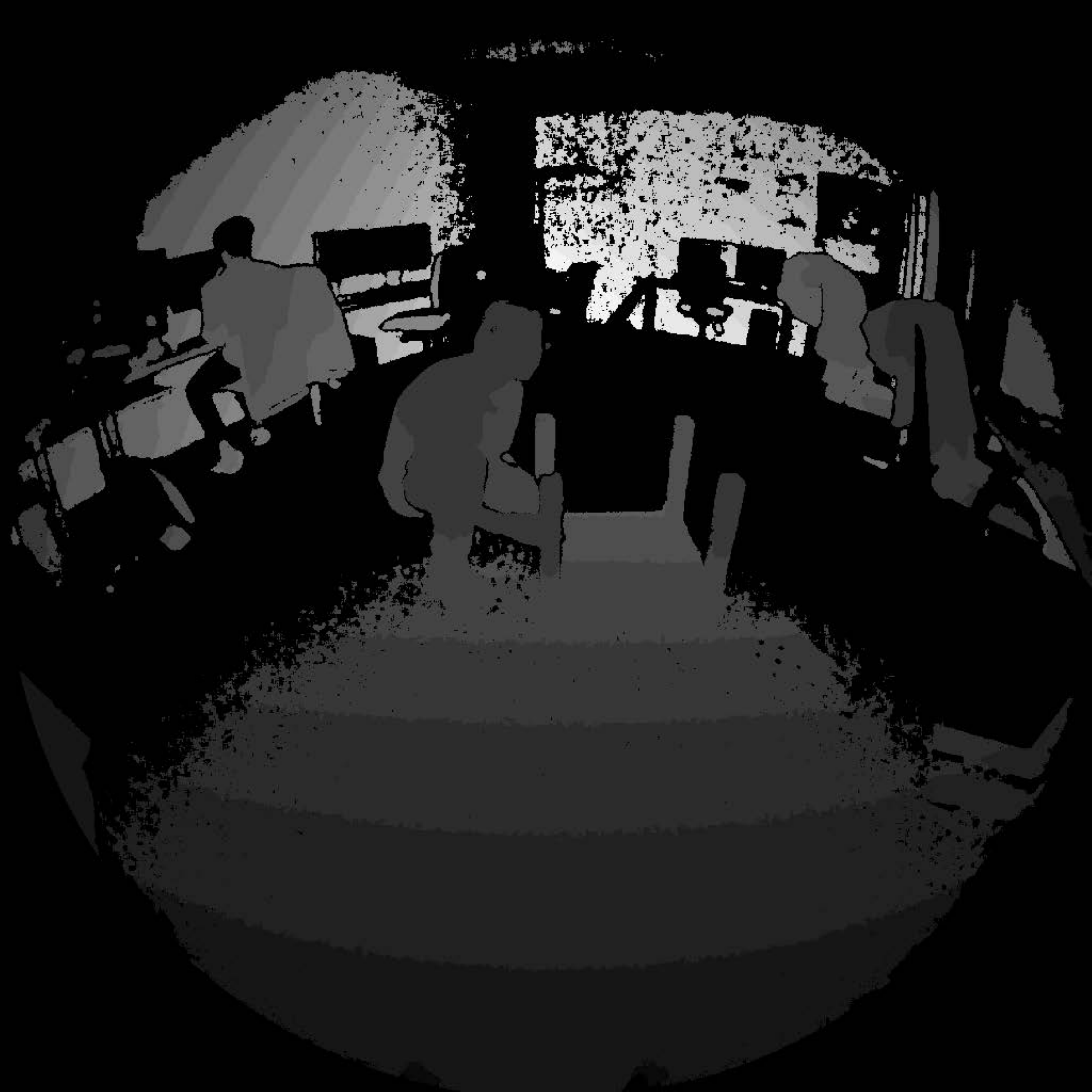}
		\end{minipage}
		\caption{Examples of images captured from one camera. The upper row are the RGB images, and we will classify the assembly action step manually; the bottom row are correspondingly normalized depth images. The shown depth images are normalized into (0,255) for display reasons.}
		\label{fig:example}
	\end{figure}

	The factory calibration of four Kinect Azure cameras is good enough for our task. They will be divided into one master device and three subordinate devices. With the help of the sensor SDK, we develop an application to start, synchronize the cameras and capture the assembly scene. Recording device data will be stored in a Matroska(.mkv) file. The Matroska stores video tracks, including colour and depth tracks. Tools such as {\it ffmpeg} or the {\it mkvinfo} command can view and extract information from recording files. Through them, we can obtain RGB images aligned with 16-bit depth images. Fig.~\ref{fig:example} shows image samples.
	
	\subsection{Data annotation}
	The fixed position of markers can help calculate the relationship between cameras, so we treat the master device's coordinate as the world coordinate. Other cameras' results will be transformed into the world coordinate.
	
	For each camera, their recorded images will be classified into different assembly states manually. For instance, the IKEA side table assembly has five states. With one more table leg being assembled, the image will be marked as the next state. After the classification, we provide the initial guess of the objects' poses only in each state's first frame to avoid manually annotating all the frames. Rest poses can be tracked through ICP algorithm.
	
	Every depth and corresponding colour frame from the four cameras with the same timestamp will be used to reconstruct the whole scene and obtain the point cloud. Then ICP algorithm will align the 3D model with the point cloud. Through the registration of multi-view data, we can receive the 6DoF object poses. Finally, all poses are refined in an optimization step.
	
	%\subsection{Statistics}
	
	\section{Conclusion and future work}
	This paper introduces the IKEA Object State Dataset, a large-scale, multi-view dataset for IKEA furniture assembly. To the best of our knowledge, most previous work pays more attention to immovable objects or understanding human behaviours. Our dataset focus on providing fully annotated frames with 6DoF pose for all object components. When hardware is available, we will continue our work on data processing and benchmark experiments.

	\clearpage
	
	\bibliographystyle{splncs04}
	\bibliography{paper}
\end{document}